# Driving Mechanisms and Forecasting of China's Pet Population: An ARIMA-RF-HW Hybrid Approach

Shengjia Chang *

College of Arts and Sciences, China University of Petroleum - Beijing at Karamay, Karamay, 834000, China, 2022016625@st.cupk.edu.cn

Xianshuo Yue

School of Information and Control Engineering, Qingdao University of Technology, Qingdao, 266520, China, 3032812048yxs@sina.com

**Abstract**

This study proposes a dynamically weighted ARIMA-RF-HW hybrid model integrating ARIMA for seasonality and trends, Random Forest for nonlinear features, and Holt-Winters smoothing for seasonal adjustment to improve China's pet population forecasting accuracy. Using 2005-2023 data with nine economic, social, and policy indicators (urban income, consumption, aging ratio, policy quantity, new veterinary drug approvals), data were preprocessed via Z-score normalization and missing value imputation. The results show that key drivers of pet populations include urban income (19.48% for cats, 17.15% for dogs), consumption (17.99% for cats), and policy quantity (13.33% for cats, 14.02% for dogs), with aging (12.81% for cats, 13.27% for dogs) and urbanization amplifying the demand for pets. Forecasts show steady cat growth and fluctuating dog numbers, reflecting cats' adaptability to urban environments. This research supports policymakers in optimizing pet health management and guides enterprises in developing differentiated services, advancing sustainable industry growth.

## 1 INTRODUCTION

China's pet industry has surged with global digitalization, reaching 300 billion yuan in 2023. Pet cats numbered 65.36 million, a 48.2% rise since 2019[1]. Key drivers include urban consumption (32,994 yuan per capita, 4.7% pet spending), aging demographics (14.9% aged 65+), 240 million singles boosting companionship demand, and tightened regulations (14 policies by 2023 vs. 5 in 2019, covering vaccines and food standards)[2].

Traditional models like ARIMA effectively capture linear trends and seasonality[3] but lack nonlinear modeling capacity. Seasonal weighting modifications improved ARIMA accuracy by 30–59%[4], though machine learning outperforms it in nonlinear contexts[5]. Machine learning techniques, particularly LSTM (achieving 1.53 RMSE in oil forecasting[6]) and Random Forests (validated in geological applications[7]), demonstrate superior nonlinear modeling capabilities. Single-model approaches remain constrained by parameter sensitivity and computational demands[8]. Ensemble methods combining linear and nonlinear components—such as ARIMA-ES-RF hybrids[9],

SVM-GA frameworks[10], and CEEMDAN-optimized systems[11] — overcome these limitations through synergistic integration. Fixed-weight integration strategies, however, show limited adaptability in dynamic environments.

This study proposes an ARIMA-RF-HW dynamic weighting framework with three innovations: (1) dynamic weight optimization via grid search to integrate ARIMA (seasonality), RF (nonlinearity), and Holt-Winters (cyclicality); (2) hybrid decomposition combining ARIMA for linear trends, RF for feature correlations, and Holt-Winters for seasonal refinement; (3) adaptive parameter calibration in Holt-Winters using prediction residuals, strengthening resilience to external disruptions (e.g., policy changes).

## 2 ARIMA - RF - HW INTEGRATED MODEL CONSTRUCTION

### 2.1 Data Preparation and Influencing Factor Analysis

Data collection is critical for analyzing key drivers of China's pet industry. Table 1 details multi-source datasets from statistical departments, including variable explanations.

Table 1. Data source.

| Data Category | Data Source |
|---|---|
| Number of Cat and Dog Owners | China Pet Development Index Report |
| Pet Ownership Penetration | China Pet Development Index Report |
| Urban Income/Spending | China Statistical Yearbook |
| Urbanization Rate | China Statistical Yearbook |
| Elderly Population/Proportion (65+) | China Statistical Yearbook |
| Single Population/Proportion (15+) | China Statistical Yearbook |
| Pet Industry Policies | Government and industry policy documents |
| New Veterinary Drug Approvals | Ministry of Agriculture and Rural Affairs reports |

The raw data contained partial missing values (approximately 2.3% of the total sample size), which were filled using linear interpolation. Z-score normalization was applied to eliminate scale differences, with the formula:

$$x' = \frac{x-\mu}{\sigma} \tag{1}$$

where $\mu$ is the feature mean and $\sigma$ is the standard deviation. After data cleaning, the distribution of all indicators passed the Kolmogorov-Smirnov test ($p > 0.05$), meeting the requirements for subsequent modeling.

Table 2 lists key indicators for China's pet industry development from prior studies[2].

Table 2. Indicators Related to the Development of China's Pet Industry.

| Primary Indicator | Secondary Indicator |
|---|---|
| Economic Environment | Urban Income |
|  | Urban Spending |
| Social Environment | Urbanization Rate |
|  | Elderly Population (65+) |
|  | Elderly Ratio (65+) |
|  | Single Population (15+) |
|  | Single Ratio (15+) |
| Policy Environment | Pet Industry Policies |
|  | New Veterinary Drugs Approved |

## 2.2 Correlation analysis

Analyze influencing factors via Spearman correlation heatmaps. Figure 1 reveals significant correlations, highlighting strong positive associations for pet cats and negative ones for dogs with key factors.

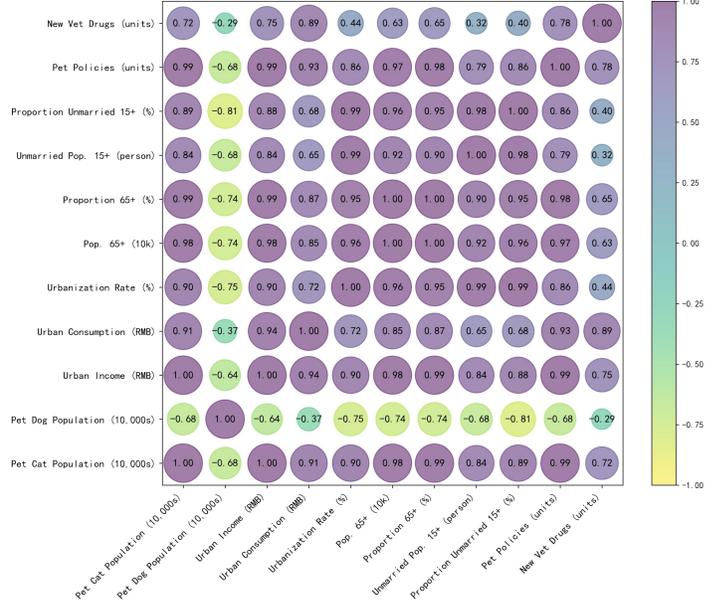

Figure 1. Factor-pet cat & dog correlation heat map.

Analysis shows multiple factors influence the pet industry. After data collection and analysis, integrated Random Forest, ARIMA, and Seasonal Seasonal Exponential Smoothing (ETS) models can address related challenges.

## 2.3 Model Architecture Design

### 2.3.1 Random Forest Model.

The importance $I(f)$ of a feature $f$ in a random forest is calculated as:

$$I(f) = \sum_{t=1}^{T}\left(ImpurityDecrease(t,f) \times \frac{N_t}{N}\right) \tag{2}$$

where $T$ is total decision trees, $ImpurityDecrease\,(t,f)$ is impurity reduction from feature $f$ in $t$-th tree, $N_t$ is samples count for splitting in $t$-th tree, and $N$ is total samples.

For a tree $t$, the impurity decrease $ImpurityDecrease(t,f)$ is:

$$ImpurityDecrease(t,f) = I_t - \left(\frac{N_{tL}}{N_t}I_{tL} + \frac{N_{tR}}{N_t}I_{tR}\right) \tag{3}$$

where $I_t$ is the impurity of the node before splitting, $I_{tL}$ and $I_{tR}$ are the impurities of the left and right child nodes after splitting, and $N_{tL}$ and $N_{tR}$ are the number of samples in the left and right child nodes, respectively[12].

Standardize feature importances $I^{'}(f)$ using:

$$I^{'}(f) = \frac{I(f) - mean(I)}{std(I)} \tag{4}$$

where mean($I$) is the mean of feature importances, and std($I$) is the standard deviation of feature importances. The flowchart of the random forest optimization algorithm is shown in Figure 2.

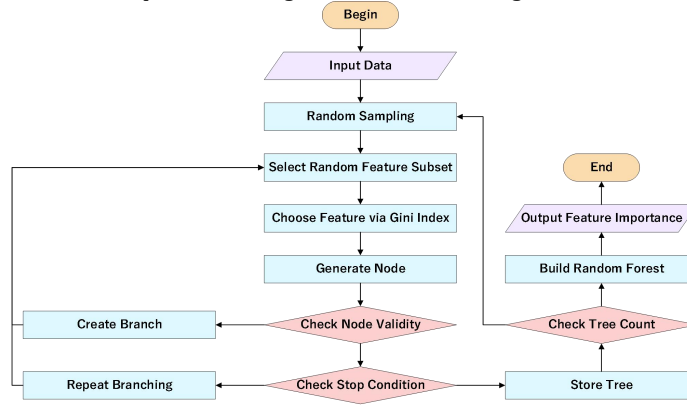

Figure 2. Random Forest Optimization Algorithm Diagram.

### 2.3.2 ARIMA Model.

ARIMA modeling begins by assessing series stationarity: build ARMA if stationary or apply differencing for ARIMA (Figure 3). Key steps: stationarity testing, identification, estimation, validation, and forecasting.

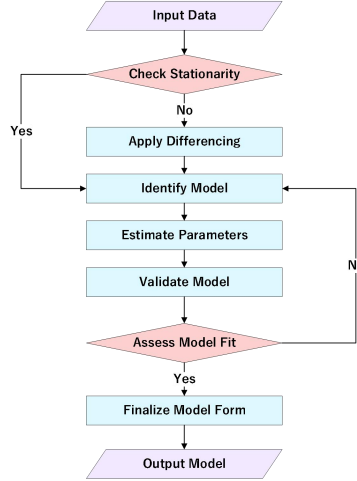

Figure 3. ARIMA Modeling and Prediction Process.

In the process of establishing an ARIMA model, the steps are as follows:

Perform a stationarity test on the time series. The commonly used test method is the Augmented Dickey-Fuller (ADF test)[13]. The regression formula of the ADF test method is as follows:

$$\nabla Y_t = \beta_0 + \lambda T + (\rho - 1)Y_{t-1} + \sum_{i=1}^{p} \gamma Y_{t-i} + \varepsilon_t \tag{5}$$

The Augmented Dickey-Fuller (ADF) test uses AIC for lag selection, testing the null hypothesis of a unit root (non-stationarity) against the stationary alternative. Non-stationarity is concluded if the test statistic exceeds critical values; otherwise, stationarity is confirmed[13]. For non-stationary series, first-order differencing is applied:

$$\nabla Y_{t'} = Y_t - Y_{t-1} = (1-B)Y_t \tag{6}$$

Therefore, the time series after D-order difference processing can be formally expressed as:

$$\nabla^d Y_t = (1-B)^d Y_t \tag{7}$$

Model identification requires selecting type and order. For stationary series, analyze autocorrelation (ACF) and partial autocorrelation (PACF); their decay patterns indicate appropriate model types and orders[14].

$$AC: \rho_k = \frac{\sum_{i=1}^{n-k} Y_i \cdot Y_{i+k}}{\sum_{i=1}^{n} Y_i^2}, k = 1,2,\cdots \tag{8}$$

$$PAC: \emptyset_{kk} = \begin{cases} \rho_1 & ,k=1 \\ \frac{\rho_k - \sum_{j=1}^{k-1} \emptyset_{k-1j} \cdot \rho_{k-j}}{1 - \sum_{j=1}^{k-1} \emptyset_{k-1j} \cdot \rho_{k-j}}, k=2,3,\cdots \end{cases} \tag{9}$$

ACF and PACF analyses determine time series model structures and initial orders through truncation patterns, supplemented by Akaike (AIC) and Schwarz (SC) criteria for parameter optimization.
The general forms of AIC and SC criteria are respectively:

$$AIC(p,q) = -\frac{2lnL}{n} + \frac{2(p+q)}{n} \tag{10}$$

$$SC(p,q) = -\frac{2lnL}{n} + \frac{2(p+q)ln(n)}{n} \tag{11}$$

In model selection, the two criteria evaluate model performance by quantifying the fit-complexity trade-off. A lower AIC or SC means a simpler structure with good fit, avoiding overfitting[14].

Following model identification, preliminary definitions of model type and order are established; parameters are then estimated using least squares to minimize residual squared sums:

$$\sum_{t=1}^{n} \varepsilon_t^2 = \sum_{t=1}^{n} (\theta_q^{-1}(B) \emptyset_p(B) \nabla^d Y_t)^2 \tag{12}$$

Validate model fit by testing residual white noise via ACF: insignificant autocorrelation coefficients confirm validity; otherwise, revise parameters.

*2.3.3 Holt-Winters Exponential Smoothing Model.*

It forecasts considering trend and seasonal components. Additive seasonal model's basic formula[15]:

$$\hat{y}_t = (L_t \cdot \alpha + T_t \cdot \beta + S_t \cdot \gamma + b_t) \tag{13}$$

In this study, we applied the Holt-Winters model to analyze time series data. Parameter settings:
1) Trend Component: "add", combined additively with the level term.
2) Seasonal Component: "add", combines with trend and level terms additively.
3) Seasonal Periods: 2, showing a repeating seasonal pattern every two time units.
Based on data analysis and preprocessing, these settings aim to capture trends and seasonal changes more accurately.

## 3 EXPERIMENTS AND RESULT ANALYSIS

Using the Random Forest model, the five important variables selected are as follows.
Top five indicators of impact on pet dogs (Table 3):

Table 3. factors affecting dog population.

| Feature | Dogs Importance |
| --- | --- |
| Urban Income (RMB) | 0.171519 |
| Pet Policies(units) | 0.140243 |
| Pop. 65+(10k) | 0.132651 |
| Urban Consumption(RMB) | 0.126789 |
| New Vet Drugs(units) | 0.096464 |

Top five indicators of impact on pet cats (Table 4):

Table 4. factors affecting cat population.

| Feature | Cats Importance |
| --- | --- |
| Urban Income (RMB) | 0.194815 |
| Urban Consumption(RMB) | 0.179860 |
| Pet Policies(units) | 0.133283 |
| Pop. 65+(10k) | 0.128112 |
| Urbanization Rate(%) | 0.106701 |

Urban income, consumption capacity, pet policies, urbanization rates, consumer spending, and aging populations drive cat/dog numbers as key pet industry factors.

ARIMA parameters (p: AR terms, d: differencing, q: MA terms) are determined through ACF/PACF plots. Due to figure volume, this section exemplifies using urban residents' disposable income (Figure 4).

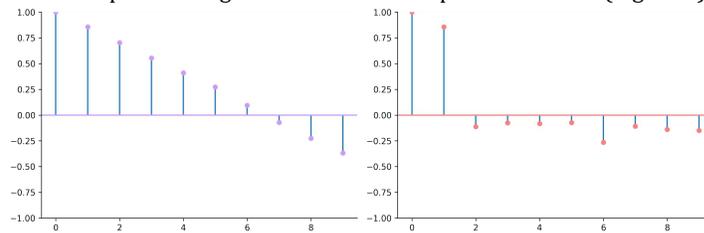

Figure 4. Per Capita Disposable Income of Urban Residents (RMB yuan) ACF and PACF.

Through analysis, for the factor "Per Capita Disposable Income of Urban Residents (RMB yuan)":
1)ACF: It shows a slow decay characteristic and has multiple significant lag values.
2)PACF: There is a significant peak at lag 1 and then decays rapidly.
3)Model Parameter Settings: $p = 1, d = 1, q = 0$.

Given that the ACF exhibits the feature of slow decay, a first-order differencing treatment is carried out to achieve the stationarity of the sequence[16].

The parameter settings for the remaining factors are obtained in a similar manner.

The predictions of important influencing indicators in the pet industry for the next three years were made using the ARIMA model, and the results are shown in Table 5:

Table 5. Predictions of Each Indicator from 2024 to 2026.

| Indicator | 2024 | 2025 | 2026 |
|---|---|---|---|
| Urban Income | 52451.76 | 52608.52 | 52647.48 |
| Urban consumption | 33393.23 | 33454.47 | 33463.89 |
| Pet policies(unite) | 13.85 | 13.70 | 13.55 |
| Pop.65(10+) | 22249 | 22721.09 | 23108.55 |
| Urbanization Rate(%) | 66.78 | 67.19 | 67.46 |
| New Vet Drugs(units) | 89.21 | 89.22 | 89.22 |

The visualization results are shown in Figure 5:

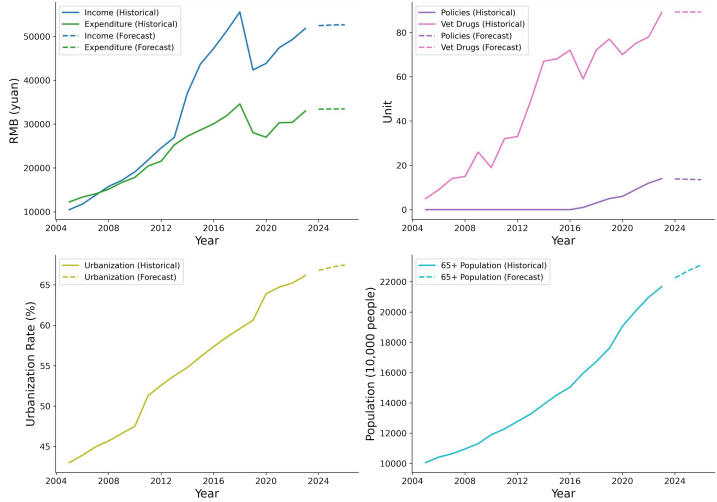

Figure 5. Predictions of Each Indicator from 2024 to 2026.

ARIMA-predicted indices inform final pet population forecasts via integrated RF and ETS models. This hybrid approach combines RF's nonlinear pattern capture (through decision trees) with ETS's seasonal trend handling, enhancing structural complexity comprehension and time-series accuracy via weighted summation:

$$\hat{y}_{combined} = \alpha \cdot \hat{y}_{rf} + \beta \cdot \hat{y}_{hw} \qquad (14)$$

Where: $\hat{y}_{combined}$ is the weighted - average prediction; $\hat{y}_{rf}$ represents the prediction result of the Random Forest model; $\hat{y}_{hw}$ represents the prediction result of the Holt-Winters Exponential Smoothing model; $\alpha$ is the weight of the Random Forest model(set as 0.7 in the code); $\beta$ the weight of the Holt-Winters model (set as 0.3 in the code). The prediction results are shown in Table 6:

Table 6. Forecasted Pet Cat and Dog Population (10,000s).

| Indicator | 2024 | 2025 | 2026 |
|---|---|---|---|
| Pet Cat Population (10,000s) | 6976 | 6751 | 6935 |
| Pet Dog Population (10,000s) | 5152 | 5189 | 5107 |

The visual result is shown in Figure 6:

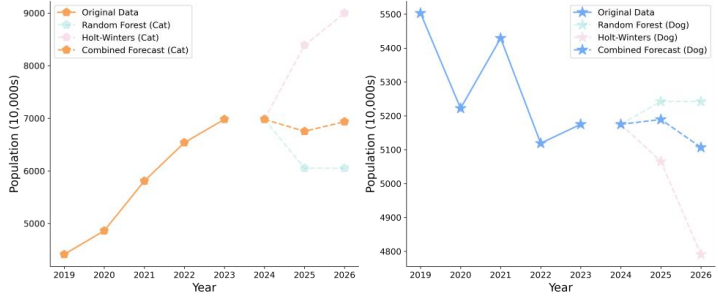

Figure 6. Forecasted Pet Cat and Dog Population (10,000s).

The number of cats shows an overall growth trend, while the number of dogs tends to stabilize or decrease after undergoing violent fluctuations. Different prediction methods have different degrees of smoothing or fitting to the changing trends of historical data, and comprehensive consideration is the best strategy.

## 4 SENSITIVITY ANALYSIS AND ERROR ANALYSIS

### 4.1 Sensitivity Analysis

Single-factor analysis and global variance decomposition (Sobol method) were employed to evaluate model sensitivity.

Single-factor analysis: By fixing other variables and adjusting urban household income by $\pm 5\%$, the predicted values fluctuated by 3%. The sensitivity coefficient was calculated as:

$$S_{income} = \frac{\Delta \hat{y}/\hat{y}_{base}}{\Delta x/x_{base}} = \frac{0.03}{0.05} = 0.6 \tag{15}$$

Global analysis: Based on Sobol index decomposition, the first-order sensitivity index for urban household income was $S_1 = 0.48$, and the total sensitivity index was $S_T = 0.62$. This indicates that income not only directly affects predictions but also amplifies effects through interactions with other variables (e.g., consumption level)[17].

### 4.2 Error analysis

Using the mean absolute error formula (MAE):

$$MAE = \frac{1}{n}\sum_{i=1}^{n}|y_i - \hat{y}_i| \tag{16}$$

Use absolute coefficient:

$$R^2 = 1 - \frac{\sum_{i=1}^{n}(y_i - \hat{y}_i)^2}{\sum_{i=1}^{n}(y_i - \bar{y}_i)^2} \tag{17}$$

Here, $y_i$ represents the actual observed values, $\hat{y}_i$ denotes the model-predicted values, $\bar{y}_i$ is the mean of observed values, and n is the number of observations[18]. The model's predicted values are generated based on the model constructed in this paper, while the actual observed values are derived from statistical data. These were then input into the Mean Absolute Error (MAE) formula and the coefficient of determination ($R^2$) to yield an MAE of 0.542 and an $R^2$ of 0.929. This demonstrates that the model's error is relatively small and within an acceptable range.

For comparison, the MAE and $R^2$ of standalone ARIMA and Random Forest models were calculated (Table 7). The ARIMA-RF-HW ensemble model significantly outperformed both ARIMA (MAE = 0.801) and Random Forest (MAE = 0.723).

Table 7. Comparison between Ensemble Models and Single Models.

| Model | MAE | RMSE | $R^2$ |
| --- | --- | --- | --- |
| ARIMA-RF-HW | 0.542 | 0.723 | 0.929 |
| ARIMA | 0.801 | 1.012 | 0.861 |
| Random Forest | 0.723 | 0.956 | 0.878 |

## 5 CONCLUSIONS AND FUTURE WORK

### 5.1 Research Conclusions

This study identifies core drivers of China's pet cat/dog population growth and mechanisms. Urban income drives 19.48% (cats) and 17.15% (dogs), the top economic factors. Consumption (17.99% for cats) and policy quantity (13.33%–14.02%) underscore consumption upgrades and policy impacts. Aging (12.81%–13.27%) and single population (15+) reflect emotional needs driving pet demand amid social changes. Forecasts show steady cat growth and slight dog fluctuations, driven by cats' urban adaptability. Declining policy volume and new veterinary drug approvals signal potential long-term constraints from policy saturation.

### 5.2 Research Prospects

Although the model demonstrates excellent performance under existing data conditions, future research could be further optimized in the following directions:
(1) Expanding data to include micro-level metrics (e.g., regional costs, breed trends) and refine regional granularity for heterogeneous urban dynamics.
(2) Enhancing models by integrating deep learning (e.g., LSTM) or adaptive ensembles to improve resilience to shocks and capture complex causal relationships.
(3) Sustainability analysis via environmental/ethical indicators (e.g., carbon footprints, stray policies) and scenario modeling for balanced industry growth and social/environmental goals.

### 5.3 Practical Significance

The findings of this study are highly consistent with the conclusion of "dual - wheel drive of consumption upgrade and policy support" proposed in the China Pet Industry Development Index Report (2023), while deepening the analysis of the impact mechanism of social aging and urbanization on the pet economy. The results provide empirical evidence for the government to optimize pet health management policies (such as vaccine standards and pet - friendly facilities in elderly communities) and for enterprises to formulate differentiated market strategies (such as upgrading pet services for high - income families), facilitating the industry's sustainable growth and rational allocation of social resources.